\pdfoutput=1
\documentclass[11pt]{article}

\usepackage[]{acl}

\usepackage{times}

\usepackage{latexsym}

\usepackage[T1]{fontenc}
\usepackage[utf8]{inputenc}

\usepackage{microtype}

\usepackage{graphicx}
\usepackage{xcolor}
\usepackage{multirow}

\usepackage{amsmath}
\usepackage{amssymb}

\title{
\vspace*{-0.5in}
{{\small \hfill COLING'2022}\\
\vspace*{.25in}} 
Dynamic Relevance Graph Network for Knowledge-Aware Question Answering}

\author{Chen Zheng \\
  Michigan State University \\
  \texttt{zhengc12@msu.edu} \\\And
  Parisa Kordjamshidi \\
  Michigan State University \\
  \texttt{kordjams@msu.edu} \\}

\begin{document}
\maketitle
\begin{abstract}
This work investigates the challenge of learning and reasoning for Commonsense Question Answering given an external source of knowledge in the form of a knowledge graph (KG). We propose a novel graph neural network architecture, called Dynamic Relevance Graph Network (DRGN). DRGN operates on a given KG subgraph based on the question and answers entities and uses the relevance scores between the nodes to establish new edges dynamically for learning node representations in the graph network. This explicit usage of relevance as graph edges has the following advantages, a) the model can exploit the existing relationships, re-scale the node weights, and influence the way the neighborhood nodes' representations are aggregated in the KG subgraph, b) It potentially recovers the missing edges in KG that are needed for reasoning. Moreover, as a byproduct, our model improves handling the negative questions due to considering the relevance between the question node and the graph entities. Our proposed approach shows competitive performance on two QA benchmarks, CommonsenseQA and OpenbookQA, compared to the state-of-the-art published results.

\end{abstract}

\section{Introduction}

Solving Question Answering (QA) problems usually requires both language understanding and human commonsense knowledge. 
Large-scale pre-trained language models (LMs) have achieved success in
many QA benchmarks~\cite{Rajpurkar2016SQuAD10,Rajpurkar2018KnowWY,Min2019MultihopRC,Yang2018HotpotQAAD}. 
However, LMs have difficulties in predicting the answer when reasoning over external knowledge is required~\cite{Yasunaga2021QAGNNRW,Feng2020ScalableMR}.

\begin{figure}[ht!]
\centering
\includegraphics[width=0.47\textwidth,height=0.25\textheight]{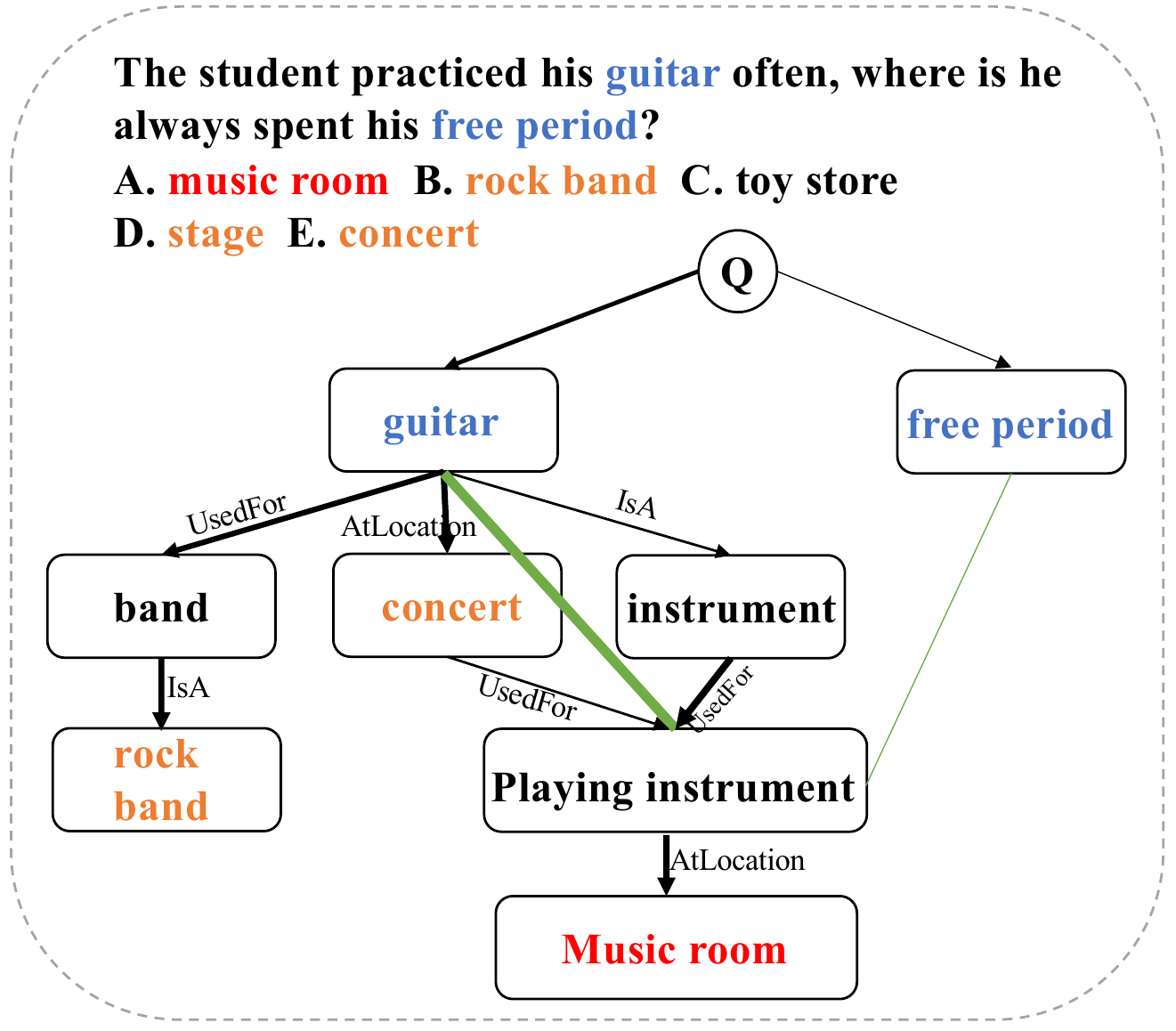}
\caption{An example of the CommonsenseQA benchmark. Given the question node Q, question entity nodes (blue boxes), correct answer entity node (red box), and wrong answer entity nodes (orange boxes), we predict the answer by reasoning over the question and the extracted KG subgraph.}
\label{fig:DRGN_csqa_example}
\end{figure}

Therefore, using the external sources of knowledge explicitly in the form of knowledge graphs (KGs) is a recent trend in question answering models~\cite{Lin2019KagNetKG,Feng2020ScalableMR}. 
Figure~\ref{fig:DRGN_csqa_example}, taken from the CommonsenseQA benchmark, shows an example for which answering the question requires commonsense reasoning. In this example, the external KG provides the required background information to obtain the reasoning chain from question to answer. 
% The current state-of-the-art models on CommonsenseQA~\cite{Yasunaga2021QAGNNRW} reason over the question and KG by adding the question node to a selected KG entity subgraph and jointly updating the representations by a graph neural model. Looking back at Figure~\ref{fig:DRGN_csqa_example}, after adding the question node to the subgraph, it becomes more clear that the student will spend the \textit{free period} in the \textit{music room} rather than the \textit{concert.}
We highlight two issues in the previous approaches taken to solve this QA problem:
a) the extracted KG subgraph sometimes misses some edges between entities, which breaks the chain of reasoning that the current models can not exploit the connections, b) the semantic context of the question and connection to the answer is not used properly, for example, reasoning when negative terms exist in the question, such as no and not, is problematic.

The first above-mentioned issue is caused by the following reasons. 
First, the knowledge graph is originally imperfect and does not include the required edges.
Second, when constructing the subgraph, to reduce the graph size, most of the models select the entities that appear in two-hop paths~\cite{Feng2020ScalableMR}.
Therefore, some intermediate concept (entity) nodes and edges are missed in the extracted KG subgraph. In such cases, the subgraph does not contain a complete chain of reasoning. Third, the current models often cannot reason over paths when there is no direct connection between the involved concepts.
While finding the chain of reasoning in QA is challenging in general~\cite{ijcai2021-553}, here this problem is more critical when the KG is the only source and there are missing edges.
% Looking back at Figure~\ref{fig:DRGN_csqa_example}, the KG subgraph misses the direct connection between ``Great Britain'' and ``island'' (green arrow), where the term ``island'' is the most crucial term in the question. 
Looking back at Figure~\ref{fig:DRGN_csqa_example}, the KG subgraph misses the direct connection between \textit{guitar} and \textit{playing instrument} (green arrow). 
For the issue of considering question semantics, as \cite{Lin2019KagNetKG} points out, previous models are not sensitive to the negation words and consequently predict opposite answers. 
QA-GNN~\cite{Yasunaga2021QAGNNRW} model is the first work to deal with the negative questions. QA-GNN improves the
reasoning under negation, to some extent, by adding the QA global node to the graph. However, the challenge still exists.

To solve the above challenges, we propose a novel architecture, called Dynamic Relevance Graph Network (DRGN).
The motivation of our proposed DRGN is to recover the missing edges and establish direct connections between concepts to facilitate multi-hop reasoning.
In particular, DRGN model uses a relational graph network module while influencing the importance of the neighbor nodes using an additional relevance matrix. 
It potentially can recover the missing edges to establish a direct connection based on the relevancy of the node representations in the KG during the training.
The module can potentially capture the connections between distant nodes while benefiting from the existing KG edges.
Our proposed model learns representations directly based on the relevance scores between subgraph entity pairs that are computed by Inner Product operation. 
At each convolutional layer of the graph neural network, we compute the inner product of the nodes based on their current layer's node representations dynamically and build the neighborhoods based on this relevance measure and form a relevance matrix accordingly.
% The neighborhoods are constructed dynamically.
This can be seen as a way to learn new edges as the training goes forward in each layer while influencing on the weights of the neighbors dynamically based on their relevance.
As shown in Figure~\ref{fig:DRGN_csqa_example}, the relevance score between \textit{guitar} and \textit{playing instrument} is stronger than other nodes in the subgraph.
Moreover, since the graph includes the question node, the relevance between the question node and entity nodes is computed at every layer, making use of the contextual information more effectively. 
% when learning node representations during the training. 
It becomes more clear that the student will spend the \textit{free period} in the \textit{music room} rather than the \textit{concert.}

In summary, the contributions of this work are as follows: 
\noindent{\bf 1)} The Proposed DRGN architecture exploits the existing edges in the KG subgraph while explicitly uses the relevance between the nodes to establish direct connections and recover the possibly missing edges dynamically. This technique helps in capturing the reasoning path in the KG for answering the question.

\noindent{\bf 2)} Our model exploits the relevance between question and the graph entities,
which helps considering the semantics of the question explicitly in the graph reasoning
and boosting the performance. In particular, it improves dealing with the negation.

\noindent{\bf 3)} Our proposed model obtains competitive results on both CommonsenseQA and OpenbookQA benchmarks. Our analysis demonstrates the significance and effectiveness of the DRGN model.

\begin{figure*}
\centering
\includegraphics[width=1.0\textwidth,height=170pt]{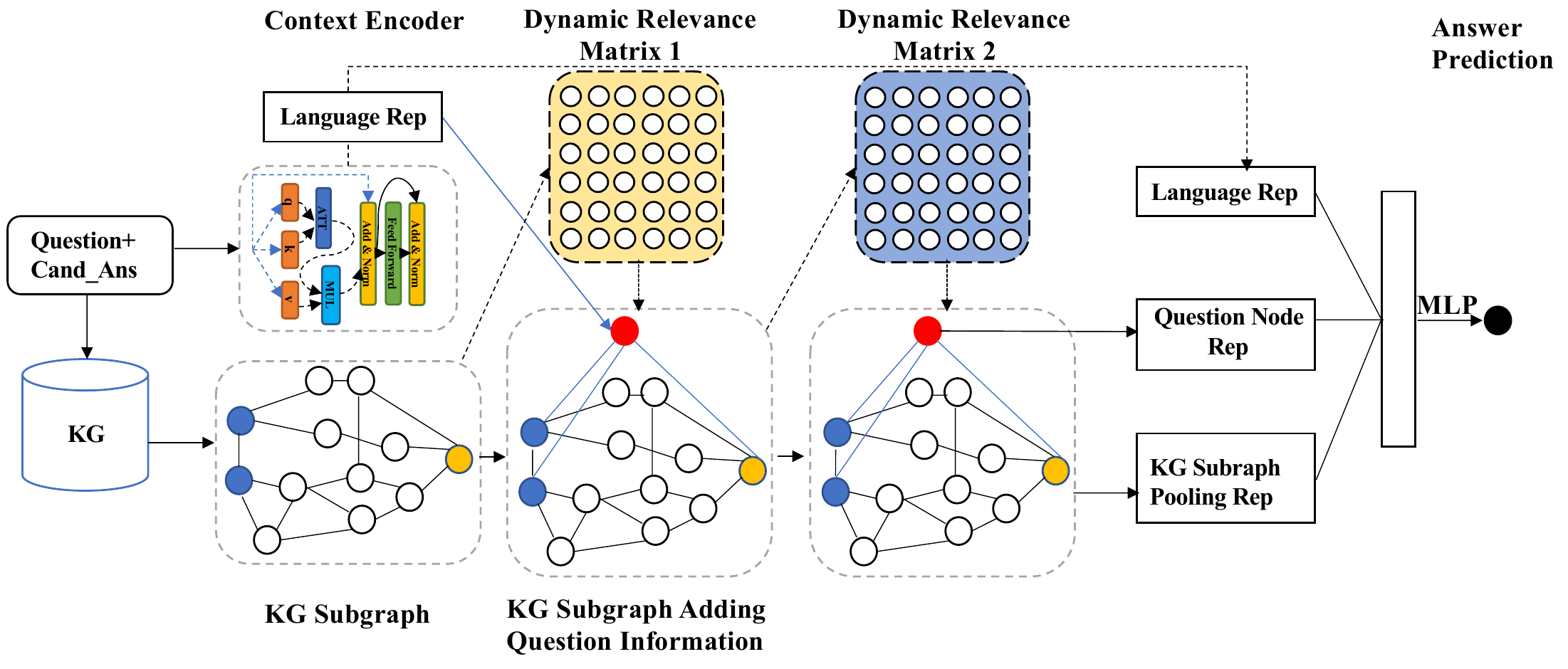}
\caption{Our proposed DRGN model is composed of the Language Context Encoder module, KG Subgraph Construction module, Graph Neural Network module, and Answer Prediction module. The blue color entity nodes represent the entity mentioned in the question. The yellow color node represents the answer node. The red color node is the question node. We use different colors to draw the dynamic relevance matrix $1$ and $2$ because the relevance matrix changes dynamically in each graph neural layer.\label{fig:architecture}}
\end{figure*}

\section{Related Work}

\subsection{QA using Knowledge Graph}
Augmenting QA systems with external knowledge has been studied in many recent research papers.
In this direction, pre-trained language models are often employed because they potentially can serve as implicit knowledge bases.~\cite{devlin-etal-2019-bert,rajaby2021time}.
To consider more interpretable knowledge,  KGs are utilized in the QA models~\cite{zheng-kordjamshidi-2022-relevant,Feng2020ScalableMR}.
However, given that the KGs are usually large and contain many nodes that are irrelevant to the question, the QA models can not effectively use the KG's information~\cite{Feng2020ScalableMR}. Moreover, with larger KGs, the computational complexity of learning over them will increase. To deal with this issue, pruning KG nodes based on a variety of metrics has been proposed~\cite{Defferrard2016ConvolutionalNN,Zhou2020GraphNN,Velickovic2018GraphAN,Hamilton2017InductiveRL,Ying2018GraphCN}.

Furthermore, the textual context is used as an additional node in the KG subgraph.
For example, \cite{KoncelKedziorski2019TextGF} and \cite{Yasunaga2021QAGNNRW} introduce the sentence node into the graph, while \cite{Fang2020HierarchicalGN} and \cite{zheng-kordjamshidi-2020-srlgrn} add the paragraph node and sentence node to construct a hierarchical graph structure. 
In this work, we also use the question node as the external node and add it to the KG subgraph. Therefore, the graph representations can learn more contextual information by computing the relevance between the question node and graph entity nodes.

\subsection{Graph Neural Networks}

Graph convolutional network (GCN)~\cite{kipf2017semi} is a classic multi-layer graph neural network. The node representations in the graph are strongly related to their neighborhood nodes and edges. For each layer of GCN, the node representations capture the information of their neighborhood nodes and edges via message passing and graph convolutional operation.
R-GCN is a variation of GCN that deals with the multi-relational graph structure~\cite{schlichtkrull2018modeling}.
\cite{Li2018AdaptiveGC} proposes an Adaptive Graph Convolution Network (AGCN) to learn the underlying relations and learn the residual graph Laplacian to improve spectral graph performance.
Meanwhile, some varients of GCN try to replace the graph Fourier transform. For example, Graph Wavelet Neural Network (GWNN)~\cite{Xu2019GraphWN} applies the graph wavelet transform to the graph, and achieves better performance compared to the graph Fourier transform in some tasks. 
% All mentioned variations of GCN operate on a given graph structure. In other words, the adjacency matrices are predefined and static matrices. 

Meanwhile, several models use the attention based Transformer operator on the graphs. 
For example, the graph attention network (GAT)~\cite{Velickovic2018GraphAN} uses the self-attention method and multi-head attention strategy to learn the node representations that consider the neighbors of the nodes. 
Besides, the gated attention network (GaAN)~\cite{Zhang2018GaANGA} uses self-attention to aggregate the different heads' information. GaAN utilizes the gate mechanism to replace the average operation that is commonly used in the GAT model.

Dynamic GCN~\cite{ye2020dynamic} is another branch of the GCN family. The dynamic graphs are constructed for different input samples. Moreover, Dynamic GCN learns the dynamic graph structure by a context-encoding network, which takes the whole feature map from the convolution neural network as input and directly predicts the adjacency matrix.
Unlike these works, our DRGN model maintains the graph structure statically, but computes the relevance edges dynamically and uses the relevance to weight the neighbors for learning node representations. Besides, our approach uses the existing relationships in the KG, recovers the missing edges and establishes the direct connections by computing the relevance between nodes dynamically. We consider this as learning new edges based on the relevancy of the nodes while the training goes forward in each layer.

% Spatial graph approaches are based on the graph topology and define convolutions on the graph~\cite{Zhou2020GraphNN}, such as DCNN~\cite{Atwood2016DiffusionConvolutionalNN} and and LGCN~\cite{Gao2018LargeScaleLG}. 
% Several models try to use the attention operator on the spatial graphs. 
% For example, the graph attention network (GAT) uses the self-attention method and multi-head attention strategy to learn the node representations that consider the neighbors of the nodes. 
% Besides, the gated attention network (GaAN)~\cite{Zhang2018GaANGA} uses self-attention to aggregate the different heads' information. GaAN utilizes the gate mechanism to replace the average operation that is commonly used in the GAT model. However, unlike our proposed dynamic relevance mechanism, the attention operator on spatial graphs maintains the local invariance and cannot capture the long-distance dependencies for the defined neighborhoods.

\section{Dynamic Relevance Graph Network}

\subsection{Task Formulation}

The task of QA over pure knowledge is to choose a correct answer $a_{ans}$ from a set of $N$ candidate answers $\{a_1, a_2,...,a_n\}$ given input question $q$ and an external knowledge graph (KG). 
In fact, the input to the problem is not the whole KG but a subgraph, $G_{sub}=(V,E)$, is selected based on previous research in~\cite{Feng2020ScalableMR} and \cite{Yasunaga2021QAGNNRW}. 
The node set $V$ represents entities in the knowledge subgraph, and the edge set $E$ represents the edges between entities.

\subsection{Model Description} 
\label{sec:model}
Figure~\ref{fig:architecture} shows the proposed Dynamic Relevance Graph Network (DRGN) architecture.
Our DRGN includes four modules: Language Context Encoder module, KG Subgraph Construction module, Graph Neural Network module, and Answer Prediction module.
% First, we concatenate the question and an answer choice to obtain the QA pair, then use the large-scale pre-trained language model (LMs) to obtain a QA pair representation (Section~\ref{sec:text_encode}). 
% Furthermore, we obtain the subgraph $G_{sub}$ from the KGs (Section~\ref{sec:graph_construct}). 
% To compute the relevance between question information and KG nodes in the dynamic relevance matrix, we add the question node as the external node to the KG subgraph that provides the additional contextual information about the question to the KG subgraph.
% This constructed graph is used as a graph neural network for learning node representations (Section~\ref{sec:dr_gnn}). 
% Finally, we make the final prediction based on the jointly learned QA pair representation and pooling representation of the subgraph (Section~\ref{sec:final_prediction}).
In this section, we describe the details of our approach and the way we train our model efficiently. 

\subsection{Language Context Encoder}
\label{sec:text_encode}
For every question $q$ and candidate answer $a_i$ pair, we concatenate them to form Language Context $L$: 
\begin{align*}
L = [ [CLS];q;[SEP]; a_i ],
\end{align*}
where [CLS] and [SEP] are the special tokens used by large-scale pre-trained Language Models (LMs).
We feed input $L$ to a pre-trained LMs encoder to obtain the list of token representations $h_L \in \mathbb{R}^{|L|* d}$, where $|L|$ represents the length of the sequence. Then we use the [CLS] representation, denoted as $h_{[CLS]} \in \mathbb{R}^d$, as the representation of $L$. 

\subsection{KG Subgraph Construction}
\label{sec:graph_construct}
%In this section, we describe how to construct the relevant commonsense subgraph.
We use ConceptNet~\cite{speer2017conceptnet}, a general-domain knowledge graph, as the commonsense KG. 
% We find the entities mentioned in the question and answer choice in the KG and extract a sub-graph that contains those entities.
ConceptNet graph has multiple semantic relational edges, e.g., HasProperty, IsA, AtLocation, etc.
We follow \cite{Feng2020ScalableMR} work to construct the subgraphs from KG for each example. The approach is to construct a subgraph from KG that contains the entities mentioned in the question and answer choices. 
The entities are selected with the exact match between n-gram tokens and ConceptNet concepts using some normalization rules. % and extract a sub-graph that contains those entities. 
Then another set of entities is added to the subgraph by following the KG paths of two hops of reasoning based on the current entities in the subgraph.

Furthermore, we add the semantic context of the question as a separate node to the subgraph.
This node provides an additional question context to the KG subgraph, $G_{sub}$, as suggested by~\cite{Yasunaga2021QAGNNRW}. 
We link the question node to entity nodes mentioned in the question. 
The semantic context of the question node $Q$ is initialized by the [CLS] representation described in Section~\ref{sec:text_encode}. The initial representation of the other entities is derived from applying RoBERTa and pooling over their contained tokens~\cite{Feng2020ScalableMR}.

\subsection{Graph Neural Network Module}
\label{sec:dr_gnn}
The basis of our learning representation is Multi-relational Graph Convolutional Network (R-GCN)~\cite{schlichtkrull2018modeling}. R-GCN is an extension of GCN that operates on a graph with multi-relational edges between nodes.
In our case, the relation types between entities are taken from the $17$ semantic relations from ConceptNet. Meanwhile, an additional type is added to represent the relationship between the question node and question entities, making the graph structure different from previous works. We denote the set of relations as $R$.

% We use $h^{(l)}$ to represent node representations in the $l$-th layer.
% In R-GCN, node representations are computed as follows:
% \begin{align}
% \label{eq:rgcn}
% h_i^{(l+1)} &= \sigma \left( W_0^{(l)}h_i^{(l)} + \sum_{r \in R}\sum_{j \in \mathcal{N}^r_i} \frac{1}{d_{i,r}}W_r^{(l)} h_j^{(l)} \right) \in \mathcal{R}^{d}, \\
% h^{(l+1)} &= [h_0^{(l+1)};h_1^{(l+1)};\cdots;h_{|V|}^{(l+1)}] \in \mathcal{R}^{|V|*d},
% \end{align}
% where $d_{i,r}$ is the normalization factor~\cite{schlichtkrull2018modeling}.
% $W_r^{(l)}$ is the learned relational weight, and $W_0$ is the learned self-loop weight, and $|V|$ is the number of nodes in the subgraph.

Our dynamic relevance graph network (DRGN) architecture is the variation of the R-GCN model. 
To establish the direct connection between the graph nodes and re-scale the importance of the neighbors, we compute the relevance score between the nodes dynamically at each graph layer based on their current learned representations. 
Then we build the neighborhoods based on this relevance measure and form a relevance matrix, $M_{rel}$, accordingly.
This can be seen as a way to learn new edges based on the relevance of the nodes as the training goes forward in each graph layer.
We use inner product to compute the relevance matrix:
\begin{align*}
\label{formu:dynamic}
M_{rel}^{(l)} = h^{(l)\top} h^{(l)} \in \mathbb{R}^{(|V|+1) * (|V|+1)},
\end{align*}
where $|V|$ is the graph entity nodes sizes, and $1$ is added due to using the question node in the graph. The relevance matrix re-scales the weights and influences the way the neighborhood nodes' representations are aggregated in the R-GCN model.
$M_{rel}$ is computed dynamically, and the relevance scores change while the representations are computed at each graph layer. 
In our proposed relational graph, the forward-pass message passing updates of the nodes, denoted by $h_i$, are calculated as follows:
\begin{align*}
h_i^{(l+1)} &= \sigma ( \sum_{r \in R}\sum_{j \in \mathbb{N}^r_i} \frac{1}{d_{i,r}}W_r^{(l)} \cdot (M_{rel_{i,j}}^{(l)} h_j^{(l)}) \\
&  \hspace{5em} + W_0^{(l)} \cdot (M_{rel_{i,i}}^{(l)} h_i^{(l)})) \in \mathbb{R}^d, 
% h_q^{(l+1)} &=  \sigma ( \sum_{j \in \mathcal{N}^q} W_q^{(l)} \cdot  F_c([h_q^{(l)};h_j^{(l)}]) + W_0^{(l)} \cdot h_q^{(l)} ),\\
% h'^{(l+1)} &= [h_0^{(l+1)};h_1^{(l+1)};\cdots;h_{|V|}^{(l+1)};h_q^{(l+1)}] \\
% &  \hspace{10em} \in \mathcal{R}^{(|V|+1)*d},
\end{align*}
% \begin{align}
% \label{formu:dynamic}
% h^{(l+1)} &= \sigma \left ( M_{rel}^{(l+1)} \cdot h'^{(l+1)} \cdot W_g \right ) \in \mathcal{R}^{(|V|+1)*d},
% \end{align}
where $\mathbb{N}^r_i$ represents the neighbor nodes of node $i$ under relation $r$, $r\in R$. $\sigma$ is the activation function, $W_r$ denotes the learnable parameters.
Besides, we calculate the updated question node representation as follows, 
\begin{align*}
h_Q^{(l+1)} &=  \sigma ( \sum_{j \in \mathbb{N}^Q} W_Q^{(l)} \cdot  F_c([h_Q^{(l)};(M_{rel_{Q,j}}^{(l)}h_j^{(l)})]) \\
&  \hspace{5em} + W_0^{(l)} \cdot (M_{rel_{Q,Q}}^{(l)} h_Q^{(l)}) )  \in \mathbb{R}^d,
\end{align*}
where $F_{c}$ is a two-layer MLP, $h_{Q}$ is the question node representation. 
Finally, we stack the node representations to form $h'^{(l+1)}$:
\begin{align*}
h'^{(l+1)} &= [h_0^{(l+1)};h_1^{(l+1)};\cdots;h_{|V|}^{(l+1)};h_Q^{(l+1)}] \\
&  \hspace{10em} \in \mathbb{R}^{(|V|+1) * d}.
\end{align*}
We then compute the $(l+1)$ layer's dynamic relevance matrix $M_{rel}^{(l+1)}$ that shows the relevance scores of node representations.
Finally, we use the $M_{rel}^{(l+1)}$ to multiply the node representation matrix $h'^{(l+1)}$ that helps the node representation to learn the weights of the edges based on the learned relevance and specifically to include the additional relevance edges between the nodes during the massage passing as follows:
\begin{align*}
h^{(l+1)} &= \sigma \left ( M_{rel}^{(l+1)} \cdot h'^{(l+1)} \cdot W_g \right ) \in \mathbb{R}^{(|V|+1) * d},
\end{align*}
where $W_g$ is the learnable parameters.

% %, $ M_{rel} \in R^{|v| \cdot |v|} $ is the dynamic relevance matrix.
% $M_{rel}(i,j)$ is a relevance score. 
% Since $M_{rel}(i,i) = 1$, the self-loop hidden representation is $W_0^{(l)} \cdot F(h_i^{(l)})$. 

%Notice that we add the question node as the external node to the KG subgraph which we mentioned in Section~\ref{sec:graph_construct}. 

% So we have an additional type of the relation between the question node and the entities. 
% We use $G_q$ to represent the question node.

% To compute the node representation that combines the text information and graph entity representation in the $l$ layer, $F(h_j^{(l)})$, we use two-layer MLP to encode the $F(h_j^{(l)})$. The process is computed as follows:
% \begin{align} 
% F(h_j^{(l)}) &= f_c(\beta_{(q,j)}^{(l)} o_j^{(l)}) \in \mathcal{R}^{d}, \\
% o_j^{(l)} &= f_{(q,j)}([h_{q}^{(l)};h_{j}^{(l)}]) \in \mathcal{R}^{d}, \\
% \beta_{(q,j)}^{(l)} &= softmax(o_j^{(l)})
% \end{align}
% where $f_{c}$ is a two-layer MLP, and $f_{(q,j)}$ is a two-layer MLP.

\subsection{Answer Prediction}
\label{sec:final_prediction}

Given the Language Context $L$ and KG subgraph, we use the information from both the language representation $h_{[CLS]}$, question node representation $h_{Q}$ learnt from the KG subgraph, and the KG subgraph representation pooled from the last graph layer,
$pool(h_{G_{sub}})$, to calculate the scores of the candidate answers as follows:
$$p(a|L, G_{sub}) = f_{out}([h_{[CLS]};h_{Q};pool(h_{G_{sub}})]),$$
where $f_{out}$ is a two-layer MLP. 
Finally, we choose the highest scored answer from $N$ candidate answers as the prediction output. We use the cross entropy loss to optimize the end-to-end model.

\section{Experiments and Results}

\subsection{Datasets}

We evaluate our model on two different QA benchmarks, CommonsenseQA and OpenbookQA. Both benchmarks come with an external knowledge graph. We apply ConceptNet to the external knowledge graph on these two benchmarks.

\noindent\textbf{CommonsenseQA}~\cite{talmor-etal-2019-commonsenseqa} is a QA dataset that requires human commonsense reasoning capacity to answer the questions. Each question in CommonsenseQA has five candidate answers without any extra information. The dataset consists of $12,102$ questions.

\noindent\textbf{OpenbookQA}~\cite{mihaylov-etal-2018-suit} It is a multiple-choice QA dataset that requires reasoning with commonsense knowledge. 
The OpenbookQA benchmark is a well-defined subset of science QA~\cite{clark2018think} that requires finding the chain of commonsense reasoning to  answer a question. Each data sample includes the question, scientific facts, and candidate answers. In our experimental setting, the scientific facts are added to the question part. This makes the problem formulation consistent with the CommonsenseQA setting.

\subsection{Implementation Details}

We implemented our DRGN architecture using PyTorch.\footnote{Our code is available at \url{https://github.com/HLR/DRGN}.} We use the pre-trained RoBERTa-large~\cite{liu2019roberta} to encode the question. We use cross-entropy loss and RAdam optimizer~\cite{Liu2020OnTV} to train our end-to-end architecture. The batch size is set to $16$, and the maximum text input sequence length set to $128$. Our model uses an early stopping strategy during the training. We use a $3$-layer graph neural module in our experiments. Section~\ref{sec:k_hops} describes the effect of the different number of layers. The learning rate for the LMs is $1e-5$, while the learning rate for the graph module is $1e-3$.

% \begin{table}[tb]
% \centering
% \small
% \scalebox{0.9}{
% \begin{tabular}{c|c}
% \hline
% \textbf{Models}& \textbf{Test}  \\
% \hline
% RoBERTa~\cite{liu2019roberta} & 72.1 \\
% STaR~\cite{zelikman2022star}&72.3\\
% HyKAS~\cite{ma-etal-2019-towards}&73.2\\
% XLNet+GraphReason~\cite{lv2020graph} & 75.3\\
% MHGRN~\cite{Feng2020ScalableMR} & 75.4  \\
% Albert~\cite{Lan2020ALBERTAL} (ensemble) &76.5          \\
% UnifiedQA (11B Parameters)~\cite{2020unifiedqa} & \textbf{79.1} \\
% QA-GNN~\cite{Yasunaga2021QAGNNRW} & 76.1 \\
% \textbf{DRGN} & \textbf{77.8} \\
% \hline
% \end{tabular}
% }\vspace{-2mm}
% \caption{CommonsenseQA official Test accuracy. }
% \vspace{-2mm}
% \label{tab:csqa_official_table}
% \end{table}

\subsection{Baseline Description}

\noindent \textbf{KagNET}~\cite{Lin2019KagNetKG} is a path-based model that models the multi-hop relations by extracting relational paths from Knowledge Graph and then encoding paths with an LSTM sequence model.

\noindent\textbf{MHGRN}~\cite{Feng2020ScalableMR}: Multi-hop Graph Relation Network (MHGRN) is a strong baseline. MHGRN model applies LMs to the question and answer context encoder, uses GNN encoder to learn graph representations, and chooses the candidate answer by these two encoders.

\noindent\textbf{QA-GNN}~\cite{Yasunaga2021QAGNNRW} is the recent SOTA model that uses a working graph to train language and KG subgraph. The model jointly reasons over the question and KG and jointly updates the representations. QA-GNN uses GAT as the backbone to do message passing on the graph. To learn the semantic edge information, QA-GNN directly adds the edge representation to the local node representation and cannot learn the global structure of the edges, which is inefficient. However, our model uses the global multi-relational adjacency matrices to learn the edge information.

\begin{table}
\begin{center}
\begin{tabular}{c|cc}
\hline
\textbf{Models}& Dev ACC$\%$ & Test ACC$\%$ \\
\hline
RoBERTa-no KG & 69.6\% & 67.8\% \\
R-GCN & 72.6\% & 68.4\% \\
GconAttn & 72.6\% & 68.5\% \\
KagNet   & 73.3\% & 69.2\% \\
RN & 73.6\% & 69.5\% \\
MHGRN & 74.4\% & 71.1\% \\
QA-GNN & 76.5\% & 73.4\% \\
\textbf{DRGN} & \textbf{78.2\%} & \textbf{74.0\%} \\
\hline
\end{tabular}
\end{center}
\caption{Dev accuracy and Test accuracy (In-House split) of various models on the CommonsenseQA benchmark, following by \cite{Lin2019KagNetKG}.}
\label{tab:csqa_acc_table}
\end{table}

\subsection{Result Comparison}
Table~\ref{tab:csqa_acc_table} shows the performance of different models on the CommonsenseQA benchmark. KagNet and MHGRN are two strong baselines. 
Our model outperforms the KagNet by $4.8\%$ and MHGRN by $2.9\%$ on CommonsenseQA benchmark. This result shows the effectiveness of our DRGN architecture.
Table~\ref{tab:obqa_acc_table} shows the performance on the OpenbookQA benchmark.
There are a few recent papers that exploit larger LMs, such as T5~\cite{JMLR:v21:20-074} that contains $3$ billion parameters ($10$x larger than our model,) and UnifiedQA~\cite{2020unifiedqa} ($32$x larger). For a fair comparison, we use the same RoBERTa setting for the input representation when we evaluate OpenbookQA. Our model performance, potentially, will be improved after using these larger LMs. To demonstrate this point, we did additional experiments using AristoRoBERTaV7~\cite{clark2019f} as a backbone to train our model. 
Our model achieves better performance when using the larger LMs compared to other baseline models. 
The performance shows that the more implicit information learned from pre-trained language models, the more effective relevance information established between graph nodes.
We should note that GREASELM~\cite{Zhang2022GreaseLMGR} and GSC~\cite{Wang2022GNNIA} are two most recent models that are developed in parallel with our DRGN. 
GREASELM aims to ground language context in commonsense knowledge graph by fusing token representations from pretrained
LMs and GNN over \textit{Modality Interaction} layers~\cite{Zhang2022GreaseLMGR}. 
GSC designs a \textit{Graph Soft Counter} layer~\cite{Wang2022GNNIA} to enhance the graph reasoning capacity.
Our results are competitive with the reported ones in those parallel works while each work emphasizes different contributions.

\begin{figure}[ht!]
\centering
\includegraphics[width=0.49\textwidth,height=140pt]{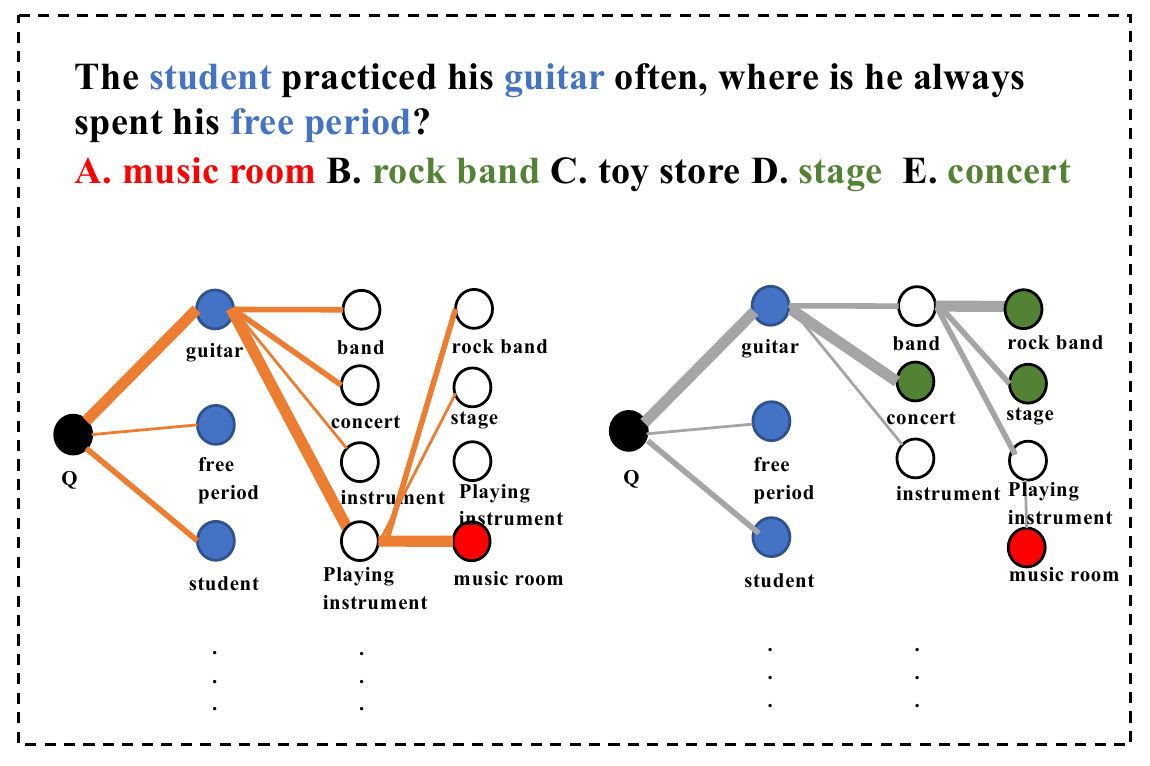}
\caption{The complete reasoning chain from the question node to the candidate answer node. The blue nodes are question entity nodes, the red and green nodes are the candidate answer nodes. The thicker edges indicate the higher relevance score to the neighborhood node, while the thinner edges indicate the lower score. The left side is the reasoning chain selected from our model (orange edges), while the right side is selected from the baseline models (grey edges).}
\label{fig:ana_example1}
\end{figure}

\begin{table}
\begin{center}\small
\begin{tabular}{c|cc}
\hline
\textbf{Models}& Dev & Test  \\
\hline
RoBERTa-large & 66.7\% & 64.8\% \\
R-GCN & 65.0\% & 62.4\% \\
GconAttn & 64.5\% & 61.9\% \\
RN & 66.8\% & 65.2\% \\
MHGRN & 68.1 \% & 66.8\% \\
QA-GNN & 68.9 \% & 67.8\% \\
% DR-GNN & 68.9 \% & 68.8\% \\
\textbf{DRGN} & \textbf{70.1\%} & \textbf{69.6\%} \\
\hline
AristoRoBERTaV7 & 79.2\% & 77.8\% \\
T5(3 Billion Parameters) & - & 83.2\% \\
UnifiedQA(11 Billion Parameters) & - & \textbf{87.2\%} \\
AristoRoBERTaV7+MHGRN & 78.6\%  & 80.6\% \\
AristoRoBERTaV7+QA-GNN & 80.4\% & 82.8\% \\
\textbf{AristoRoBERTaV7+DRGN} & \textbf{81.8\%} & \textbf{84.1\%} \\
\hline
\end{tabular}
\end{center}
\caption{Development and Test accuracy of various model performance on the OpenbookQA benchmark.}
\label{tab:obqa_acc_table}
\end{table}

\section{Analysis}

\subsection{Effects on Finding the Line of Reasoning}

In this section, we analyze the effectiveness of our DRGN model that helps in recovering the missing edges and establishing direct connections based on the relevancy of the node representations in the KG.
As we described in Section~\ref{sec:graph_construct}, to keep the graph size small, most of the models construct the KG subgraph via selecting the entities that appear in two-hop paths.
Therefore, some intermediate concept nodes and edges are missed in the extracted KG subgraph, and the complete reasoning chain from the question entity node to the candidate answer node can not be found.

For example, as shown in Figure~\ref{fig:ana_example1}, the question is ``The student practiced his guitar often, where is he always spent his free period?'' and the answer is ``music room''. The reasoning chain includes $2$ hops, that is, ``guitar $\rightarrow$ playing instrument $\rightarrow$ music room''. 
Since the constructed graph misses the direct edge between ``guitar'' and ``playing instrument'', MHGRN and QA-GNN baselines select the wrong intermediate node and predict the wrong answer ``concert'' and ``rock band'' by the grey edges described in the Figure~\ref{fig:ana_example1}. In contrast, our DRGN model makes a correct prediction by computing the relevance score of the nodes based on their learned representations and forming new edges accordingly. 
As we describe in Section~\ref{sec:graph_construct}, our model initializes the entity node representation by large-scale pre-trained language models (LMs). The implicit representations of LMs are learned from the huge corpora, and the knowledge is implicitly learned. 
Therefore, these two entities, ``guitar'' and ``playing instrument'', start with an implicit connection. By looking at the relevance changes, after several layers of graph encoding, the relevance score between ``guitar'' and ``playing instrument'' becomes stronger. In contrast, the relevance score between ``guitar'' and ``concert'' becomes weaker because of the contextual information ``free period''. This is the primary reason why our DRGN model obtains the correct reasoning chain.

% Looking into another example, the question is \textit{``Bob spent all of his time analyzing the works of masters because he wanted to become what?''} and the answer is ``intelligent''. The reasoning chain is ``analyze $\rightarrow$ work $\rightarrow$ master $\rightarrow$ learn $\rightarrow$ intelligent''. 
% Due to the missing link between ``master $\rightarrow$ learn'' in KG, the baseline model, MHGRN, cannot obtain the complete reasoning chain. However, our DRGN captures the strong relevance between ``master'' and ``learn'' and finds out the complete reasoning chain to predict the correct answer.

\begin{figure}
\centering
\includegraphics[width=0.49\textwidth,height=130pt]{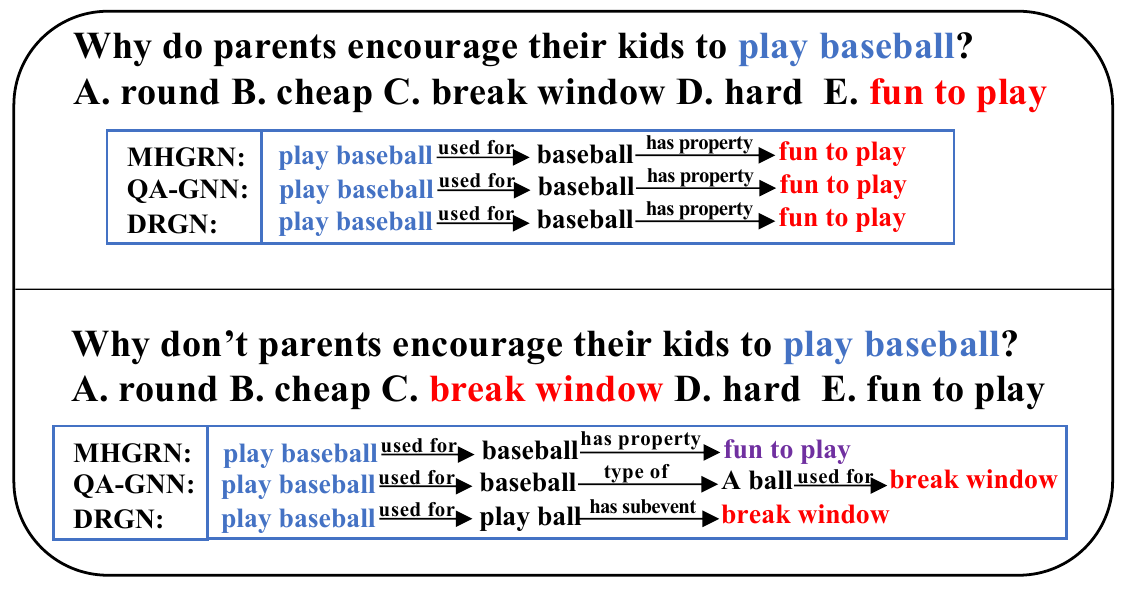}
\caption{The case study of the negation examples. The question in the bottom box includes the negation words. The red colored text represents the gold answer, and the purple colored represents the wrong answer. In the blue box, each line represents the commonsense reasoning chain of each model.}
\label{fig:negation}
\end{figure}

\begin{table}
\begin{center}\small
\begin{tabular}{c|cc}
\hline
\textbf{Models}&  Test ACC $\%$ & Test ACC$\%$ \\
&                 (Overall) & question w/ negative \\
\hline
RoBERTa-large & 68.7 \% & 54.2\% \\
KagNet & 69.2 \% & 54.2 \% \\
MHGRN & 71.1 \% & 54.8\% \\
QA-GNN & 73.4 \% & 58.8\% \\
DRGN & \textbf{75.0\%} & \textbf{60.1\%} \\
\hline
\end{tabular}
\end{center}
\caption{Performance on questions with negation in In-house split test CommonsenseQA.}
\label{tab:full_negat}
\end{table}

\subsection{Effects on Semantic Context}
While the graph has a broad coverage of knowledge, the semantic context of the question and connection to the answer is not used properly. For example, dealing with negation can not perform well~\cite{Yasunaga2021QAGNNRW}. 
Since our dynamic relevance matrix includes the semantic context of the question, the relevance between the question and graph entities is computed at every graph neural layer while considering the negation in the node representations. Intuitively, this should improve handling the negative question in our model.

To analyze this hypothesis for DRGN architecture, we compare the performance of various models on questions containing negative
words (e.g., no, not, nothing, unlikely) from CommonsenseQA following \cite{Yasunaga2021QAGNNRW}. 
The result is shown in Table~\ref{tab:full_negat}.
We observe that the baseline models of KagNet and MHGRN provide limited improvements over RoBERTa on questions with negation words (+0.4\%). However, our DRGN model exhibits a huge boost (+5.9\%).
Moreover, the DRGN model gains a larger improvement in the accuracy compared to the QA-GNN model, demonstrating the effectiveness of considering relevance between question semantics and graph entity that experimentally confirms our hypothesis. 
An additional ablation study in Table~\ref{tab:ablation} confirms this idea further. When removing the question information from DRGN, we observe that the performance on negation becomes close to the MHGRN.

Figure~\ref{fig:negation} shows qualitative examples about the positive and negative questions. For the positive question, all the models obtain the same reasoning chain ``play baseball-(used for)$\rightarrow$ baseball-(has property)$\rightarrow$ fun to play'', including MHGRN, QA-GNN, and our architecture.
However, when adding the negative words, MHGRN obtains the same reasoning chain as the positive situation, while QA-GNN and DRGN find the correct reasoning chain. One interesting finding is that DRGN can detect the direct connection using fewer hops to establish the reasoning chain. 
% The chain of reasoning is found by looking at the neighborhood matrices of the models. 
% For DRGN, this neighborhood is the outcome of multiplication of neighborhood matrix by the relevance matrix. 

\begin{figure}
\centering
\includegraphics[width=0.45\textwidth,height=110pt]{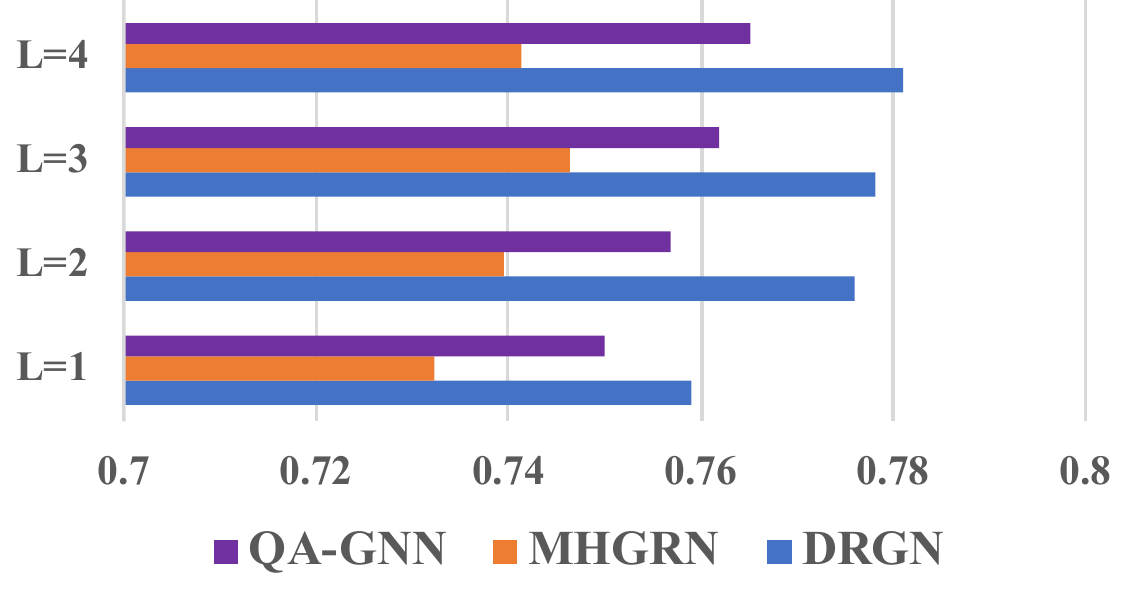}
\caption{The Effect of number of layers in QA-GNN, MHGRN, and DRGN models on CommonsenseQA.}
\label{fig:hops}
\end{figure}

\begin{table}
\begin{center}\small
\begin{tabular}{l|cc}
\hline
\textbf{Models} &  Time &  Space \\
\hline
$l$-layer KagNet & $O(|R|^l|V|^{l+1}l)$ & $O(|R|^l|V|^{l+1}l)$  \\
$l$-layer MHGRN &  $O(|R|^2|V|^2l)$  & $O(|R||V|l)$  \\
$l$-layer QA-GNN & $O(|V|^2l)$ & $O(|R||V|l)$   \\
$l$-layer DRGN &   $O(|R|^2|V|^2l)$ & $O(|R||V|^2l)$   \\
\hline
\end{tabular}
\end{center}
\caption{The time complexity and space complexity comparison between DRGN and baseline models.}
\label{tab:complexity}
\end{table}

\subsection{Effects of Number of Graph Layers}
\label{sec:k_hops}
The number of graph layers is an influencing factor for DRGN architecture because our relevance matrix is computed dynamically, and the relevance scores change while the representations are computed at each graph layer. 
We evaluate the effects of multiple layers $l$ for the baseline models and our DRGN by evaluating its performance on the CommonsenseQA. As shown in Figure~\ref{fig:hops}, the increase of $l$ continues to bring benefits until $l = 4$ for DRGN. We compare the performance after adding each layer for MHGRN, QA-GNN, and our DRGN. We observe that DRGN consistently achieves the best performance with the same number of layers as the baselines.

Table~\ref{tab:complexity} shows the time complexity and the space complexity comparison between DRGN model and baseline model. We compare the computational complexity based on the number of layers $l$, the number of nodes $V$, and the number of relations $R$. 
% We report this for the baseline models based on ~\citeauthor{Yasunaga2021QAGNNRW} work. 
Our model and MHGRN have the same time complexity because both models use the R-GCN model as the backbone. 
Besides, QA-GNN directly adds the edge representation to the local node representation during the graph pre-processing step and learns the graph node representation without the global semantic relational adjacency matrices. 
% the scope of the GAT attention is still in the predefined neighborhood area in the KG subgraph. However, in our DRGN model, we consider all nodes in the subgraph.
% It is hard to evaluate which of the two models, R-GCN and GAT, is more effective as a backbone of our setting. However, 
After adding the dynamic relevance matrix at each graph layer, our DRGN model achieves better performance compared to other baseline architectures.
For the space complexity, our model's space complexity is slightly larger than MHGRN because DRGN introduces the extra dynamic relevance matrix. However, this cost depends on the size of the subgraph, which is usually small while it leads to a huge improvement.

\begin{table}
\begin{center}
\begin{tabular}{l|cc}
\hline
\textbf{Models} &  Dev ACC \\
\hline
DRGN w/o KG subgraph & 69.6\%  \\
+ KG subgraph & 72.6\%  \\
+ relational edges in graph & 73.7\%   \\
+ question node in graph & 74.9\%  \\
+ dynamic relevance matrix & 78.2\%   \\
\hline
% DRGN graph dim = 100 & 77.3\%  \\
% DRGN graph dim = 200 & \textbf{78.1\%} \\
% DRGN graph dim = 300 & 77.8 \%   \\
% \hline
\end{tabular}
\end{center}
\caption{Ablation Study on CommonsenseQA dataset.}
\label{tab:ablation}
\end{table}

\subsection{Ablation of DRGN Modules}

To evaluate the effectiveness of various components of DRGN, we perform an ablation study on the CommonsenseQA development benchmark. Table~\ref{tab:ablation} shows the results of ablation study. 
First, we remove the whole commonsense subgraph. Our model without the subgraph obtains $69.6\%$ on the CommonsenseQA. This shows how the implicit language model can answer the questions without the external KG, which is not high-performing but yet impressive. After adding the KG subgraph, the accuracy improves to $72.6\%$ on the CommonsenseQA benchmark. 
Second, we keep the KG subgraph and add multiple relational edge information from the subgraph (described in section~\ref{sec:dr_gnn}). Without the relational edges, the accuracy becomes $73.7\%$. This result shows that the multiple relational edges help in learning better graph node representations and obtaining a higher performance.
Third, we keep the multi-relational subgraph and add the question node. In other words, we incorporate the semantic relationship between the question node and the graph entities.
The accuracy of the model improves to $74.9\%$. 
% It demonstrates the importance of the relevance between the question information and the KG subgraph.
Finally, we add the most important component, the dynamic relevance matrix, to each graph layer. The  large improvement demonstrates the importance of the dynamic relevance matrix and the effectiveness of  DRGN architecture.

\section{Conclusion}
In this paper, we propose a novel Dynamic Relevance Graph Network (DRGN) architecture for commonsense question answering given an external source of knowledge in the form of a Knowledge Graph. Our model learns the graph node representation while a) exploits the existing relations in KG, b) re-scales the importance of the neighbor nodes in the graph based on training a dynamic relevance matrix, c) establishes direct connections between graph nodes based on measuring the relevance scores of the nodes dynamically during training. The dynamic relevance edges help in finding the chain of reasoning when there are missing edges in the original KG. Our quantitative and qualitative analysis shows that the proposed approach facilitates answering the complex questions that need multiple hops of reasoning. Furthermore, since DRGN uses the relevance between the question node and graph entities, it exploits the richer semantic context of the question in graph reasoning which leads to improvements in the performance on the negative questions.
Our proposed approach shows competitive performance on two QA benchmarks, including CommonsenseQA and OpenbookQA.

\section*{Acknowledgments}
We thank all reviewers for their suggestions and helpful comments. This project is supported by National Science Foundation (NSF) CAREER award $\#$2028626 and partially supported by the Office of Naval Research (ONR) grant $\#$N00014-20-1-2005.
Any opinions, findings, and conclusions or recommendations expressed in this material are those of the authors and do not necessarily reflect the views of the National Science Foundation nor the Office of Naval Research.
\bibliography{acl2022}
\bibliographystyle{acl_natbib}

\end{document}